\documentclass[10pt,twocolumn,letterpaper]{article}

\usepackage{iccv}
\usepackage{times}
\usepackage{epsfig}
\usepackage{graphicx}
\usepackage{amsmath}
\usepackage{amssymb}

\usepackage{multirow}
\usepackage{makecell}
\usepackage{tabulary}

\usepackage{stackengine}
\newcommand\oast{\stackMath\mathbin{\stackinset{c}{0ex}{c}{0ex}{\ast}{\bigcirc}}}

\usepackage[pagebackref=true,breaklinks=true,letterpaper=true,colorlinks,bookmarks=false]{hyperref}

\iccvfinalcopy 

\ificcvfinal\fi

\begin{document}

\title{Contextual Transformer Networks for Visual Recognition}

\author{Yehao Li, Ting Yao, Yingwei Pan, and Tao Mei \\
{\normalsize\centering JD AI Research, Beijing, China}\\
{\tt\small \{yehaoli.sysu, tingyao.ustc, panyw.ustc\}@gmail.com, tmei@jd.com}
}

\maketitle
\ificcvfinal\thispagestyle{empty}\fi

\begin{abstract}
   Transformer with self-attention has led to the revolutionizing of natural language processing field, and recently inspires the emergence of Transformer-style architecture design with competitive results in numerous computer vision tasks. Nevertheless, most of existing designs directly employ self-attention over a 2D feature map to obtain the attention matrix based on pairs of isolated queries and keys at each spatial location, but leave the rich contexts among neighbor keys under-exploited. In this work, we design a novel Transformer-style module, i.e., Contextual Transformer (\textbf{CoT}) block, for visual recognition. Such design fully capitalizes on the contextual information among input keys to guide the learning of dynamic attention matrix and thus strengthens the capacity of visual representation. Technically, CoT block first contextually encodes input keys via a $3\times3$ convolution, leading to a static contextual representation of inputs. We further concatenate the encoded keys with input queries to learn the dynamic multi-head attention matrix through two consecutive $1\times1$ convolutions. The learnt attention matrix is multiplied by input values to achieve the dynamic contextual representation of inputs. The fusion of the static and dynamic contextual representations are finally taken as outputs. Our CoT block is appealing in the view that it can readily replace each $3\times3$ convolution in ResNet architectures, yielding a Transformer-style backbone named as Contextual Transformer Networks (\textbf{CoTNet}). Through extensive experiments over a wide range of applications (e.g., image recognition, object detection and instance segmentation), we validate the superiority of CoTNet as a stronger backbone.
   Source code is available at \url{https://github.com/JDAI-CV/CoTNet}.
\end{abstract}

\section{Introduction}

\begin{figure}[!tb]
\vspace{-0.33in}
\centering {\includegraphics[width=0.46\textwidth]{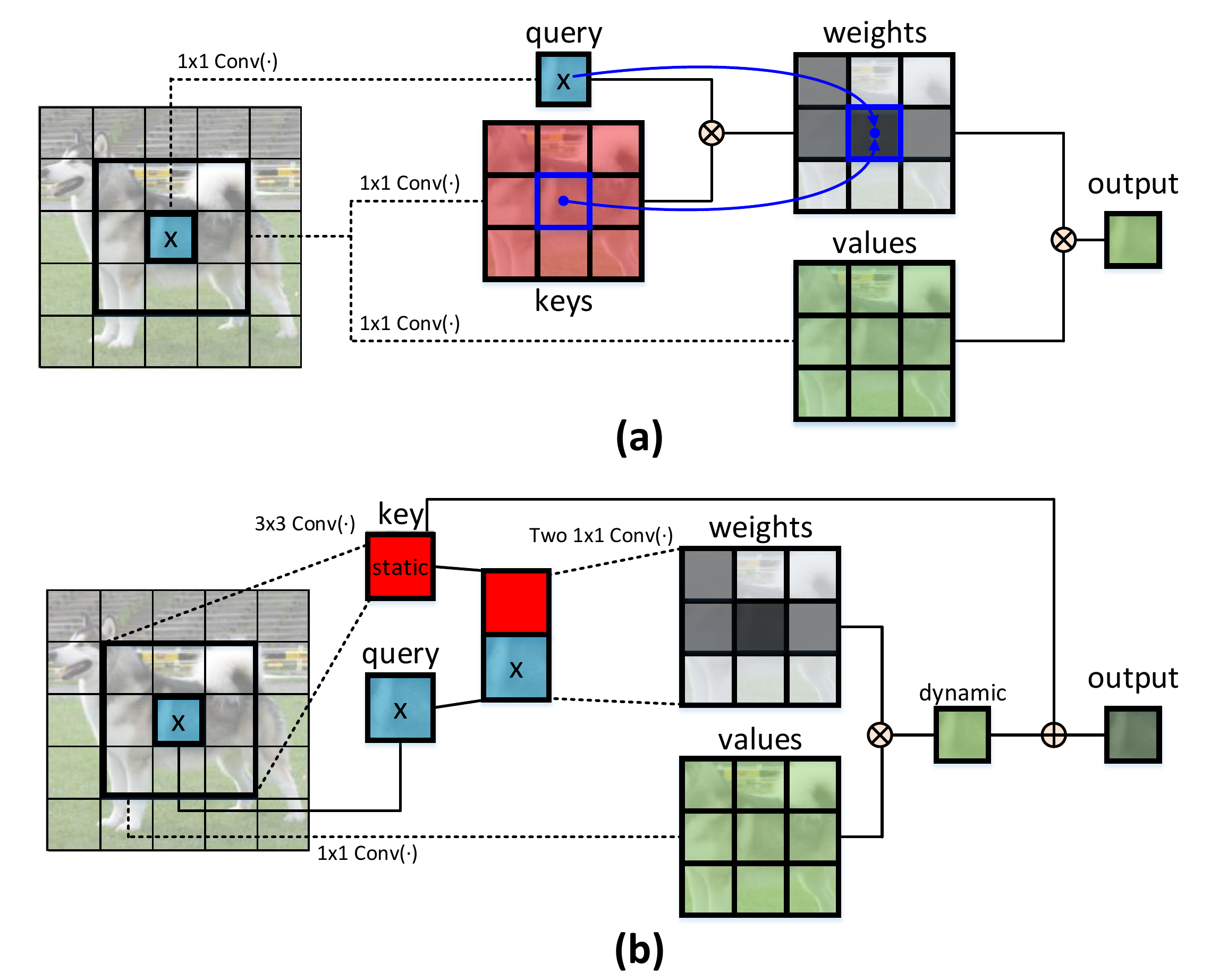}}
\vspace{-0.1in}
\caption{Comparison between conventional self-attention and our Contextual Transformer (CoT) block. (a) Conventional self-attention solely exploits the isolated query-key pairs to measure attention matrix, but leaves rich contexts among keys under-exploited. Instead, (b) CoT block first mines the static context among keys via a 3$\times$3 convolution. Next, based on the query and contextualized key, two consecutive 1$\times$1 convolutions are utilized to perform self-attention, yielding the dynamic context. The static and dynamic contexts are finally fused as outputs.}
\label{fig:fig1}
\vspace{-0.28in}
\end{figure}

Convolutional Neural Networks (CNN) \cite{chollet2017xception,dai2017deformable,he2016deep,krizhevsky2012imagenet,simonyan2014very,szegedy2015going,tan2019efficientnet} demonstrates high capability of learning discriminative visual representations, and convincingly generalizes well to a series of Computer Vision (CV) tasks, e.g., image recognition, object detection, and semantic segmentation. The de-facto recipe of CNN architecture design is based on discrete convolutional operators (e.g., 3$\times$3 or 5$\times$5 convolution), which effectively impose spatial locality and translation equivariance. However, the limited receptive field of convolution adversely hinders the modeling of global/long-range dependencies, and such long-range interaction subserves numerous CV tasks \cite{mottaghi2014role,rabinovich2007objects}. Recently, Natural Language Processing (NLP) field has witnessed the rise of Transformer with self-attention in powerful language modeling architectures \cite{devlin2018bert,vaswani2017attention} that triggers long-range interaction in a scalable manner. Inspired by this, there has been a steady momentum of breakthroughs \cite{bello2019attention,carion2020end,dosovitskiy2020image,li2021scheduled,pan2020x,ramachandran2019stand,zhao2020exploring} that push the limits of CV tasks by integrating CNN-based architecture with Transformer-style modules. For example, ViT \cite{dosovitskiy2020image} and DETR \cite{carion2020end} directly process the image patches or CNN outputs using self-attention as in Transformer. \cite{ramachandran2019stand,zhao2020exploring} present a stand-alone design of local self-attention module, which can completely replace the spatial convolutions in ResNet architectures.
Nevertheless, previous designs mainly hinge on the independent pairwise query-key interaction for measuring attention matrix as in conventional self-attention block (Figure \ref{fig:fig1} (a)), thereby ignoring the rich contexts among neighbor~keys.

In this work, we ask a simple question - \emph{is there an elegant way to enhance Transformer-style architecture by exploiting the richness of context among input keys over 2D feature map}? For this purpose, we present a unique design of Transformer-style block, named Contextual Transformer (CoT), as shown in Figure \ref{fig:fig1} (b). Such design unifies both context mining among keys and self-attention learning over 2D feature map in a single architecture, and thus avoids introducing additional branch for context mining. Technically, in CoT block, we first contextualize the representation of keys by performing a 3$\times$3 convolution over all the neighbor keys within the 3$\times$3 grid. The contextualized key feature can be treated as a \emph{static} representation of inputs, that reflects the \emph{static} context among local neighbors. After that, we feed the concatenation of the contextualized key feature and input query into two consecutive $1\times1$ convolutions, aiming to produce the attention matrix. This process naturally exploits the mutual relations among each query and all keys for self-attention learning with the guidance of the \emph{static} context. The learnt attention matrix is further utilized to aggregate all the input values, and thus achieves the \emph{dynamic} contextual representation of inputs to depict the \emph{dynamic} context. We take the combination of the \emph{static} and \emph{dynamic} contextual representation as the final output of CoT block. In summary, our launching point is to simultaneously capture the above two kinds of spatial contexts among input keys, i.e., the \emph{static} context via 3$\times$3 convolution and the \emph{dynamic} context based on contextualized self-attention, to boost visual representation learning.

Our CoT can be viewed as a unified building block, and is an alternative to standard convolutions in existing ResNet architectures without increasing the parameter and FLOP budgets.
By directly replacing each 3$\times$3 convolution in a ResNet structure with CoT block, we present a new Contextual Transformer Networks (dubbed as CoTNet) for image representation learning.
Through extensive experiments over a series of CV tasks, we demonstrate that our CoTNet outperforms several state-of-the-art backbones.
Notably, for image recognition on ImageNet, CoTNet obtains a 0.9\% absolute reduce of the top-1 error rate against ResNeSt (101 layers). For object detection and instance segmentation on COCO, CoTNet absolutely improves ResNeSt with 1.5\% and 0.7\% mAP, respectively.

\section{Related Work}

\subsection{Convolutional Networks} Sparked by the breakthrough performance on ImageNet dataset via AlexNet \cite{krizhevsky2012imagenet}, Convolutional Networks (ConvNet) has become a dominant architecture in CV field. One mainstream of ConvNet design follows the primary rule in LeNet \cite{lecun1998gradient}, i.e., stacking low-to-high convolutions in series by going deeper: 8-layer AlexNet, 16-layer VGG \cite{simonyan2014very}, 22-layer GoogleNet \cite{szegedy2015going}, and 152-layer ResNet \cite{he2016deep}. After that, a series of innovations have been proposed for ConvNet architecture design to strengthen the capacity of visual representation. For example, inspired by split-transform-merge strategy in Inception modules, ResNeXt \cite{xie2017aggregated} upgrades ResNet with aggregated residual transformations in the same topology.
DenseNet \cite{huang2017densely} additionally enables the cross-layer connections to boost the capacity of ConvNet.
Instead of exploiting spatial dependencies in ConvNet \cite{jaderberg2015spatial,mottaghi2014role}, SENet \cite{hu2018squeeze,hu2020squeeze} captures the interdependencies between channels to perform channel-wise feature recalibration.
\cite{tan2019efficientnet} further scales up an auto-searched ConvNet to obtain a family of EfficientNet networks, which achieve superior accuracy and~efficiency.

\subsection{Self-attention in Vision} Taking the inspiration from self-attention in Transformer that continuously achieves the impressive performances in various NLP tasks, the research community starts to pay more attention to self-attention in vision scenario. The original self-attention mechanism in NLP domain \cite{vaswani2017attention} is devised to capture long-range dependency in sequence modeling. In vision domain, a simple migration of self-attention mechanism from NLP to CV is to directly perform self-attention over feature vectors across different spatial locations within an image. In particular, one of the early attempts of exploring self-attention in ConvNet is the non-local operation \cite{wang2018non} that severs as an additional building block to employ self-attention over the outputs of convolutions. \cite{bello2019attention} further augments convolutional operators with global multi-head self-attention mechanism to facilitate image classification and object detection. Instead of using global self-attention over the whole feature map \cite{bello2019attention,wang2018non} that scale poorly, \cite{hu2019local,ramachandran2019stand,zhao2020exploring} employ self-attention within local patch (e.g., 3$\times$3 grid). Such design of local self-attention effectively limits the parameter and computation consumed by the network, and thus can fully replace convolutions across the entirety of deep architecture. Recently, by reshaping raw images into a 1D sequence, a sequence Transformer \cite{chen2020generative} is adopted to auto-regressively predict pixels for self-supervised representation learning. Next, \cite{carion2020end,dosovitskiy2020image} directly apply a pure Transformer to the sequences of local features or image patches for object detection and image recognition. Most recently, \cite{srinivas2021bottleneck} designs a powerful backbone by replacing the final three 3$\times$3 convolutions in a ResNet with global self-attention layers.

\begin{figure*}[!tb]
\vspace{-0.1in}
    \centering {\includegraphics[width=0.9\textwidth]{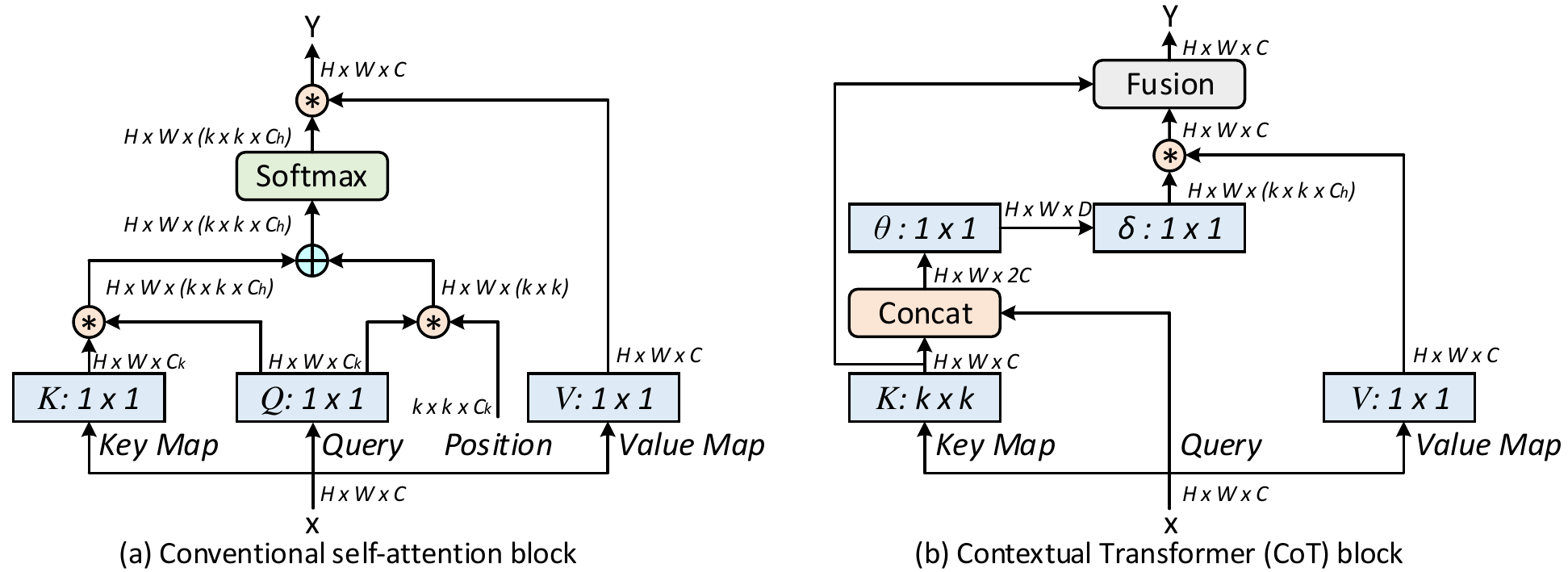}}
    \vspace{-0.05in}
    \caption{The detailed structures of (a) conventional self-attention block and (b) our Contextual Transformer (CoT) block. \textcircled{+} and  \textcircled{$\ast$} denotes the element-wise sum and local matrix multiplication, respectively.}
    \label{fig:framework}
    \vspace{-0.2in}
\end{figure*}

\subsection{Summary} Here we also focus on exploring self-attention for the architecture design of vision backbone. Most of existing techniques directly capitalize on the conventional self-attention and thus ignore the explicit modeling of rich contexts among neighbor keys. In contrast, our Contextual Transformer block unifies both context mining among keys and self-attention learning over feature map in a single architecture with favorable parameter budget.

\section{Our Approach}

In this section, we first provide a brief review of the conventional self-attention widely adopted in vision backbones. Next, a novel Transformer-style building block, named Contextual Transformer (CoT), is introduced for image representation learning. This design goes beyond conventional self-attention mechanism by additionally exploiting the contextual information among input keys to facilitate self-attention learning, and finally improves the representational properties of deep networks. After replacing 3$\times$3 convolutions with CoT block across the whole deep architecture, two kinds of Contextual Transformer Networks, i.e., CoTNet and CoTNeXt deriving from ResNet \cite{he2016deep} and ResNeXt \cite{xie2017aggregated}, respectively, are further elaborated.

\subsection{Multi-head Self-attention in Vision Backbones}

Here we present a general formulation for the scalable local multi-head self-attention in vision backbones \cite{hu2019local,ramachandran2019stand,zhao2020exploring}, as depicted in Figure \ref{fig:framework} (a). Formally, given an input 2D feature map $X$ with the size of $H \times W \times C$ ($H$: height, $W$: width, $C$: channel number), we transform $X$ into queries $Q = XW_q$, keys $K = XW_k$, and values $V=XW_v$ via embedding matrix ($W_q$, $W_k$, $W_v$), respectively. Notably, each embedding matrix is implemented as 1$\times$1 convolution in space. After that, we obtain the local relation matrix $R \in {\mathbb{R}}^{{H}\times{W}\times{(k \times k \times C_h)}}$ between keys $K$ and queries $Q$ as:
\begin{equation}\small
\label{eq:sa1}
R=K \oast Q,
\end{equation}
where $C_h$ is the head number, and $\oast$ denotes the local matrix multiplication operation that measures the pairwise relations between each query and the corresponding keys within the local $k \times k$ grid in space. Thus, each feature $R^{(i)}$ at $i$-th spatial location of $R$ is a $k \times k \times C_h$-dimensional vector, that consists of $C_h$ local query-key relation maps (size: $k \times k$) for all heads. The local relation matrix $R$ is further enriched with the position information of each $k \times k$ grid:
\begin{equation}\small
\label{eq:sa2}
\hat{R}=R+P \oast Q,
\end{equation}
where $P\in {\mathbb{R}}^{k \times k \times C_k}$ represents the 2D relative position embeddings within each $k \times k$ grid, and is shared across all $C_h$ heads.
Next, the attention matrix $A$ is achieved by normalizing the enhanced spatial-aware local relation matrix $\hat{R}$ with Softmax operation along channel dimension for each head: $A=\text{\texttt{Softmax}}(\hat{R})$. After reshaping the feature vector at each spatial location of $A$ into $C_h$ local attention matrices (size: $k \times k$), the final output feature map is calculated as the aggregation of all values within each $k \times k$ grid with the learnt local attention matrix:
\begin{equation}\small
\label{eq:sa3}
Y=V \oast A.
\end{equation}
Note that the local attention matrix of each head is only utilized for aggregating evenly divided feature map of $V$ along channel dimension, and the final output $Y$ is the concatenation of aggregated feature maps for all heads.

\subsection{Contextual Transformer Block}

Conventional self-attention nicely triggers the feature interactions across different spatial locations depending on the inputs themselves. Nevertheless, in the conventional self-attention mechanism, all the pairwise query-key relations are independently learnt over isolated query-key pairs, without exploring the rich contexts in between. That severely limits the capacity of self-attention learning over 2D feature map for visual representation learning.
To alleviate this issue, we construct a new Transformer-style building block, i.e., Contextual Transformer (CoT) block in Figure \ref{fig:framework} (b), that integrates both contextual information mining and self-attention learning into a unified architecture. Our launching point is to fully exploit the contextual information among neighbour keys to boost self-attention learning in an efficient manner, and strengthen the representative capacity of the output aggregated feature map.

\begin{table}[!tb]\scriptsize
  \centering
  \caption{\small The detailed structures of ResNet-50 (left) and CoTNet-50 (right). The shapes and operations within a residual building block are shown inside the brackets and the number of stacked blocks in each stage is listed outside. CoTNet-50 has a slightly smaller number of parameters and FLOPs than ResNet-50.}
  \setlength\extrarowheight{1.1pt}
\begin{tabular}{c|c|c|c}
\Xhline{2\arrayrulewidth}
stage      & ResNet-50                                  & \textbf{CoTNet-50}        & \!\!output\!\!    \\ \hline
res1       & 7 $\!\times\!$ 7 conv, 64, stride 2  & 7 $\!\times\!$ 7 conv, 64, stride 2    & \!\!112 $\!\times\!$ 112\!\!\\ \hline
\multirow{2}{*}{res2} & 3 $\!\times\!$ 3 max pool, stride 2  & 3 $\!\times\!$ 3 max pool, stride 2 & \multirow{2}{*}{\!\!56 $\!\times\!$ 56\!\!}\\ \cline{2-3}
                      &  $\left[ \begin{array}{l} 1 \!\times\! 1,64\\ 3 \!\times\! 3,64\\ 1 \!\times\! 1,256 \end{array} \right] \!\times\! 3$
                  &  $\left[ \begin{array}{l} 1 \!\times\! 1,64\\ {\textbf{\color{blue}CoT}},64\\ 1 \!\times\! 1,256 \end{array} \right] \!\times\! 3$
                  &      \\ \hline

res3
       & $\left[ \begin{array}{l} 1 \!\times\! 1,128\\ 3 \!\times\! 3,128\\ 1 \!\times\! 1,512 \end{array} \right] \!\times\! 4$
       & $\left[ \begin{array}{l} 1 \!\times\! 1,128\\ {\textbf{\color{blue}CoT}},128\\ 1 \!\times\! 1,512 \end{array} \right] \!\times\! 4$
       & \!\!28 $\!\times\!$ 28\!\! \\ \hline
res4
       & $\left[ \begin{array}{l} 1 \!\times\! 1,256\\ 3 \!\times\! 3,256\\ 1 \!\times\! 1,1024 \end{array} \right] \!\times\! 6$
       & $\left[ \begin{array}{l} 1 \!\times\! 1,256\\ {\textbf{\color{blue}CoT}},256\\ 1 \!\times\! 1,1024 \end{array} \right] \!\times\! 6$
       & \!\!14 $\!\times\!$ 14\!\! \\ \hline
res5
       & $\left[ \begin{array}{l} 1 \!\times\! 1,512\\ 3 \!\times\! 3,512\\ 1 \!\times\! 1,2048 \end{array} \right] \!\times\! 3$
       & $\left[ \begin{array}{l} 1 \!\times\! 1,512\\ {\textbf{\color{blue}CoT}},512\\ 1 \!\times\! 1,2048 \end{array} \right] \!\times\! 3$
       & \!\!7 $\!\times\!$ 7\!\! \\ \hline
 & \makecell{global average pool \\ 1000-d fc, softmax} & \makecell{global average pool \\ 1000-d fc, softmax} & 1 $\!\times\!$ 1 \\ \hline
\# params                     & \textbf{25.56} $\!\times\!$ $10^6$        & \textbf{22.21} $\!\times\!$ $10^6$  &   \\ \hline
FLOPs                         & \textbf{4.12} $\!\times\!$ $10^9$     & \textbf{3.28} $\!\times\!$ $10^9$       &   \\ \Xhline{2\arrayrulewidth}
\end{tabular}
\vspace{-0.2in}
\label{table:ResNet}
\end{table}

In particular, suppose we have the same input 2D feature map $X \in {\mathbb{R}}^{H \times W \times C}$. The keys, queries, and values are defined as $K=X$, $Q=X$, and $V=XW_v$, respectively. Instead of encoding each key via 1$\times$1 convolution as in typical self-attention, CoT block first employs $k \times k$ group convolution over all the neighbor keys within $k \times k$ grid spatially for contextualizing each key representation. The learnt contextualized keys $K^1\in {\mathbb{R}}^{H \times W \times C}$ naturally reflect the static contextual information among local neighbor keys, and we take $K^1$ as the static context representation of input $X$. After that, conditioned on the concatenation of contextualized keys $K^1$ and queries $Q$, the attention matrix is achieved through two consecutive 1$\times$1 convolutions ($W_{\theta}$ with ReLU activation function and $W_{\delta}$ without activation function):
\begin{equation}\small
\label{eq:cot1}
A=[K^1, Q]W_{\theta}W_{\delta}.
\end{equation}
In other words, for each head, the local attention matrix at each spatial location of $A$ is learnt based on the query feature and the contextualized key feature, rather than the isolated query-key pairs. Such way enhances self-attention learning with the additional guidance of the mined static context $K^1$. Next, depending on the contextualized attention matrix $A$, we calculate the attended feature map $K^2$ by aggregating all values $V$ as in typical self-attention:
\begin{equation}\small
\label{eq:cot2}
K^2=V \oast A.
\end{equation}
In view that the attended feature map $K^2$ captures the dynamic feature interactions among inputs, we name $K^2$ as the dynamic contextual representation of inputs. The final output of our CoT block ($Y$) is thus measured as the fusion of the static context $K^1$ and dynamic context $K^2$ through attention mechanism \cite{li2019selective}.

\begin{table}[!tb]\scriptsize
  \centering
  \caption{\small The detailed structures of ResNeXt-50 with a 32$\times$4d template (left) and CoTNeXt-50 with a 2$\times$48d template (right). The shapes and operations within a residual building block are shown inside the brackets and the number of stacked blocks in each stage is listed outside. $C$ denotes the number of groups within grouped convolutions.
  Compared to ResNeXt-50, CoTNeXt-50 has a slightly larger number of parameters but similar FLOPs.
}
  \setlength\extrarowheight{1.1pt}
\begin{tabular}{c|c|c|c}
\Xhline{2\arrayrulewidth}
\!\!stage      & ResNeXt-50 (32$\!\times\!$4d)        & \textbf{CoTNeXt-50 (2$\!\times\!$48d)}  & \!\!output\!\!      \\ \hline
\!\!res1       & 7 $\!\times\!$ 7 conv, 64, stride 2  & 7 $\!\times\!$ 7 conv, 64, stride 2 & \!\!112 $\!\times\!$ 112\!\!  \\ \hline
\multirow{2}{*}{\!\!res2}
              & 3 $\!\times\!$ 3 max pool, stride 2  & 3 $\!\times\!$ 3 max pool, stride 2 & \multirow{2}{*}{\!\!56 $\!\times\!$ 56\!\!} \\ \cline{2-3}
     &  \!$\left[\!\! \begin{array}{l} 1 \!\times\! 1,128\\ 3 \!\times\! 3,128,C\!\!=\!\!32\\ 1 \!\times\! 1,256 \end{array} \!\!\right] \!\!\times\!\! 3$\!\!
     &  \!$\left[\!\! \begin{array}{l} 1 \!\times\! 1,96\\ {\textbf{\color{blue}CoT}},96,C\!\!=\!\!2\\ 1 \!\times\! 1,256 \end{array} \!\!\right] \!\!\times\!\! 3$\!\!
     & \\ \hline

\!\!res3
     & \!$\left[\!\! \begin{array}{l} 1 \!\times\! 1,256\\ 3 \!\times\! 3,256,C\!\!=\!\!32\\ 1 \!\times\! 1,512 \end{array} \!\!\right] \!\!\times\!\! 4$\!\!
     & \!$\left[\!\! \begin{array}{l} 1 \!\times\! 1,192\\ {\textbf{\color{blue}CoT}},192,C\!\!=\!\!2\\ 1 \!\times\! 1,512 \end{array} \!\!\right] \!\!\times\!\! 4$\!\!
     & \!\!28 $\!\times\!$ 28\!\! \\ \hline
\!\!res4
     & \!$\left[\!\! \begin{array}{l} 1 \!\times\! 1,512\\ 3 \!\times\! 3,512,C\!\!=\!\!32\\ 1 \!\times\! 1,1024 \end{array} \!\!\right] \!\!\times\!\! 6$\!\!
     & \!$\left[\!\! \begin{array}{l} 1 \!\times\! 1,384\\ {\textbf{\color{blue}CoT}},384,C\!\!=\!\!2\\ 1 \!\times\! 1,1024 \end{array} \!\!\right] \!\!\times\!\! 6$\!\!
     & \!\!14 $\!\times\!$ 14\!\! \\ \hline
\!\!res5
     & \!$\left[\!\! \begin{array}{l} 1 \!\times\! 1,1024\\ 3 \!\times\! 3,1024,C\!\!=\!\!32\\ 1 \!\times\! 1,2048 \end{array} \!\!\right] \!\!\times\!\! 3$\!\!
     & \!$\left[\!\! \begin{array}{l} 1 \!\times\! 1,768\\ {\textbf{\color{blue}CoT}},768,C\!\!=\!\!2\\ 1 \!\times\! 1,2048 \end{array} \!\!\right] \!\!\times\!\! 3$\!\!
     & \!\!7 $\!\times\!$ 7\!\! \\ \hline
     & \makecell{global average pool \\ 1000-d fc, softmax} & \makecell{global average pool \\ 1000-d fc, softmax} & \!\!1 $\!\times\!$ 1\!\! \\ \hline
\# params                     & \textbf{25.03} $\!\times\!$ $10^6$        & \textbf{30.05} $\!\times\!$ $10^6$  &   \\ \hline
FLOPs                         & \textbf{4.27} $\!\times\!$ $10^9$     & \textbf{4.33} $\!\times\!$ $10^9$       &   \\ \Xhline{2\arrayrulewidth}
\end{tabular}
\vspace{-0.2in}
\label{table:ResNeXt}
\end{table}

\subsection{Contextual Transformer Networks}

The design of our CoT is a unified self-attention building block, and acts as an alternative to standard convolutions in ConvNet. As a result, it is feasible to replace convolutions with their CoT counterparts for strengthening vision backbones with contextualized self-attention. Here we present how to integrate CoT blocks into existing state-of-the-art ResNet architectures (e.g., ResNet \cite{he2016deep} and ResNeXt \cite{xie2017aggregated}) without increasing parameter budget significantly. Table \ref{table:ResNet} and Table \ref{table:ResNeXt} shows two different constructions of our Contextual Transformer Networks (CoTNet) based on the ResNet-50/ResNeXt-50 backbone, called CoTNet-50 and CoTNeXt-50, respectively. Please note that our CoTNet is flexible to generalize to deeper networks (e.g., ResNet-101).

\textbf{CoTNet-50.} Specifically, CoTNet-50 is built by directly replacing all the 3$\times$3 convolutions (in the stages of res2, res3, res4, and res5) in ResNet-50 with CoT blocks. As our CoT blocks are computationally similar with the typical convolutions, CoTNet-50 has similar (even slightly smaller) parameter number and FLOPs with ResNet-50.

\textbf{CoTNeXt-50.} Similarly, for the construction of CoTNeXt-50, we first replace all the 3$\times$3 convolution kernels in group convolutions of ResNeXt-50 with CoT blocks. Compared to typical convolutions, the depth of the kernels within group convolutions is significantly decreased when the number of groups (i.e., $C$ in Table \ref{table:ResNeXt}) is increased. In ResNeXt-50, the computational cost of group convolutions is thus reduced by a factor of $C$. Therefore, in order to achieve the similar parameter number and FLOPs with ResNeXt-50, we additionally reduce the scale of input feature map of CoTNeXt-50 from 32$\times$4d to 2$\times$48d. Finally, CoTNeXt-50 requires only 1.2$\times$ more parameters and 1.01$\times$ more FLOPs than ResNeXt-50.

\subsection{Connections with Previous Vision Backbones}

In this section, we discuss the detailed relations and differences between our Contextual Transformer and the previous most related vision backbones.

\textbf{Blueprint Separable Convolution \cite{haase2020rethinking}} approximates the conventional convolution with a 1$\times$1 pointwise convolution plus a $k \times k$ depthwise convolution, aiming to reduce the redundancies along depth axis. In general, such design has some commonalities with the transformer-style block (e.g., the typical self-attention and our CoT block). This is due to that the transformer-style block also utilizes 1$\times$1 pointwise convolution to transform the inputs into values, and the followed aggregation computation with $k \times k$ local attention matrix is performed in a similar depthwise manner. Besides, for each head, the aggregation computation in transformer-style block adopts channel sharing strategy for efficient implementation without any significant accuracy drop. Here the utilized channel sharing strategy can also be interpreted as the tied block convolution \cite{wang2020tied}, which shares the same filters over equal blocks of channels.

\textbf{Dynamic Region-Aware Convolution \cite{chen2020dynamic}} introduces a filter generator module (consisting of two consecutive 1$\times$1) to learn specialized filters for region features at different spatial locations. It therefore shares a similar spirit with the attention matrix generator in our CoT block that achieves dynamic local attention matrix for each spatial location. Nevertheless, the filter generator module in \cite{chen2020dynamic} produces the specialized filters based on the primary input feature map. In contrast, our attention matrix generator fully exploits the complex feature interactions between contextualized keys and queries for self-attention learning.

\textbf{Bottleneck Transformer \cite{srinivas2021bottleneck}} is the contemporary work, which also aims to augment ConvNet with self-attention mechanism by replacing 3$\times$3 convolution with Transformer-style module. Specifically, it adopts global multi-head self-attention layers, which are computationally more expensive than local self-attention in our CoT block. Therefore, with regard to the same ResNet backbone, BoT50 in \cite{srinivas2021bottleneck} only replaces the final three 3$\times$3 convolutions with Bottleneck Transformer blocks, while our CoT block can completely replace 3$\times$3 convolutions across the whole deep architecture. In addition, our CoT block goes beyond typical local self-attention in \cite{hu2019local,ramachandran2019stand,zhao2020exploring} by exploiting the rich contexts among input keys to strengthen self-attention learning.

\section{Experiments}
In this section, we verify and analyze the effectiveness of our Contextual Transformer Networks (CoTNet) as a backbone via empirical evaluations over multiple mainstream CV applications, ranging from image recognition, object detection, to instance segmentation. Specifically, we first undertake experiments for image recognition task on ImageNet benchmark \cite{deng2009imagenet} by training our CoTNet from scratch. Next, after pre-training CoTNet on ImageNet, we further evaluate the generalization capability of the pre-trained CoTNet when transferred to downstream tasks of object detection and instance segmentation on COCO dataset \cite{lin2014microsoft}.

\subsection{Image Recognition}

\textbf{Setup.} We conduct image recognition task on the ImageNet dataset, which consists of 1.28 million training images and 50,000 validation images derived from 1,000 classes. Both of the top-1 and top-5 accuracies on the validation set are reported for evaluation. For this task, we adopt two different training setups in the experiments, i.e., the default training setup and advanced training setup.

The default training setup is the widely adopted setting in classic vision backbones (e.g., ResNet \cite{he2016deep}, ResNeXt \cite{xie2017aggregated}, and SENet \cite{hu2018squeeze}), that trains networks for around 100 epochs with standard preprocessing. Specifically, each input image is cropped into 224$\times$224, and only the standard data augmentation (i.e., random crops and horizontal flip with 50\% probability) is performed. All the hyperparameters are set as in official implementations without any additional tuning. Similarly, our CoTNet is trained in an end-to-end manner, through backpropagation using SGD with momentum 0.9 and label smoothing 0.1. We set the batch size as $B=512$ that enables applicable implementations on an 8-GPU machine. For the first five epochs, the learning rate is scaled linearly from 0 to $\frac{0.1 \cdot B}{256}$, which is further decayed via cosine schedule \cite{loshchilov2016sgdr}. As in \cite{bello2021lambdanetworks}, we adopt exponential moving average with weight 0.9999 during training.

For fair comparison with state-of-the-art backbones (e.g., ResNeSt \cite{zhang2020resnest}, EfficientNet \cite{tan2019efficientnet} and LambdaNetworks \cite{bello2021lambdanetworks}), we additionally involve the advanced training setup with longer training epochs and improved data augmentation \& regularization. In this setup, we train our CoTNet with 350 epochs, coupled with the additional data augmentation of RandAugment \cite{cubuk2020randaugment} and mixup \cite{zhang2017mixup}, and the regularization of dropout \cite{srivastava2014dropout} and DropConnect \cite{wan2013regularization}.

\textbf{Performance Comparison.} We compare with several state-of-the-art vision backbones with two different training settings (i.e., default and advanced training setups) on ImageNet dataset. The performance comparisons are summarized in Tables \ref{table:r1} and \ref{table:r2} for each kind of training setup, respectively. Note that we construct several variants of our CoTNet and CoTNeXt with two kinds of depthes (i.e., 50-layer and 101-layer), yielding CoTNet-50/101 and CoTNeXt-50/101. In advanced training setup, as in LambdaResNet \cite{bello2021lambdanetworks}, we additionally include an upgraded version of our CoTNet, i.e., SE-CoTNetD-101, where the 3$\times$3 convolutions in the res4 and res5 stages are replaced with CoT blocks under SE-ResNetD-50 \cite{he2019bag,bello2021revisiting} backbone. Moreover, in default training setup, we also report the performances of our models with the use of exponential moving average for fair comparison against LambdaResNet.

\begin{table}[!tb]\small
  \centering
  \caption{Performance comparisons with the state-of-the-art vision backbones for image recognition on ImageNet (default training setup). Models with same depth (50-layer/101-layer) are grouped for efficiency comparison. $^\star$ indicates the use of exponential moving average during training.}
  \setlength{\tabcolsep}{0.5pt}
\begin{tabular}{c|c|cc|cc}
\Xhline{2\arrayrulewidth}
Backbone                                                & Res. & Params & GFLOPs & Top-1 Acc. & Top-5 Acc. \\ \hline
ResNet-50 \cite{he2016deep}                             & 224  & 25.5M  & 4.1    & 77.3       & 93.6       \\
Res2Net-50 \cite{gao2019res2net}                        & 224  & 25.7M  & 4.3    & 78.0       & 93.9       \\
ResNeXt-50 \cite{xie2017aggregated}                     & 224  & 25.0M  & 4.2    & 78.2       & 93.9       \\
SE-ResNeXt-50 \cite{hu2018squeeze}                      & 224  & 27.6M  & 4.3    & 78.6       & 94.2       \\
LR-Net-50 \cite{hu2019local}                            & 224  & 23.3M  & 4.3    & 77.3       & 93.6          \\
Stand-Alone$^\star$ \cite{ramachandran2019stand}        & 224  & 18.0M  & 3.6    & 77.6       & -          \\
AA-ResNet-50 \cite{bello2019attention}                  & 224  & 25.8M  & 4.2    & 77.7       & 93.8       \\
BoTNet-S1-50 \cite{srinivas2021bottleneck}              & 224  & 20.8M  & 4.3    & 77.7       & -          \\
ViT-B/16 \cite{dosovitskiy2020image}                    & 384  & -      & -      & 77.9       & -            \\
SAN19  \cite{zhao2020exploring}                         & 224  & 20.5M  & 3.3    & 78.2       & 93.9       \\
LambdaResNet-50$^\star$\cite{bello2021lambdanetworks}   & 224  & 15.0M  & -      & 78.4       & -          \\ \hline
\textbf{CoTNet-50}                                      & 224  & 22.2M  & 3.3    & \textbf{79.2}  & \textbf{94.5}  \\
\textbf{CoTNet-50}$^\star$                              & 224  & 22.2M  & 3.3    & \textbf{79.8}  & \textbf{94.9}  \\
\textbf{CoTNeXt-50}                                     & 224  & 30.1M  & 4.3    & \textbf{79.5}  & \textbf{94.5}  \\
\textbf{CoTNeXt-50}$^\star$                             & 224  & 30.1M  & 4.3    & \textbf{80.2}  & \textbf{95.1}  \\
\textbf{SE-CoTNetD-50}                                  & 224  & 23.1M  & 4.1    & \textbf{79.8}  & \textbf{94.7}  \\
\textbf{SE-CoTNetD-50}$^\star$                          & 224  & 23.1M  & 4.1    & \textbf{80.5}  & \textbf{95.2}  \\\hline\hline
ResNet-101 \cite{he2016deep}                            & 224  & 44.6M  & 7.9    & 78.5       & 94.2       \\
ResNeXt-101 \cite{xie2017aggregated}                    & 224  & 44.2M  & 8.0    & 79.1       & 94.4       \\
Res2Net-101 \cite{gao2019res2net}                       & 224  & 45.2M  & 8.1    & 79.2       & 94.4       \\
SE-ResNeXt-101 \cite{hu2018squeeze}                     & 224  & 49.0M  & 8.0    & 79.4       & 94.6        \\
LR-Net-101  \cite{hu2019local}                          & 224  & 42.0M  & 8.0    & 78.5       & 94.3       \\
AA-ResNet-101 \cite{bello2019attention}                 & 224  & 45.4M  & 8.1    & 78.7       & 94.4       \\ \hline
\textbf{CoTNet-101}                                     & 224  & 38.3M  & 6.1    & \textbf{80.0}  & \textbf{94.9}  \\
\textbf{CoTNet-101}$^\star$                             & 224  & 38.3M  & 6.1    & \textbf{80.9}  & \textbf{95.3}  \\
\textbf{CoTNeXt-101}                                    & 224  & 53.4M  & 8.2    & \textbf{80.3}  & \textbf{95.0}  \\
\textbf{CoTNeXt-101}$^\star$                            & 224  & 53.4M  & 8.2    & \textbf{81.3}  & \textbf{95.6}  \\
\textbf{SE-CoTNetD-101}                                 & 224  & 40.9M  & 8.5    & \textbf{80.5}  & \textbf{95.1}  \\
\textbf{SE-CoTNetD-101}$^\star$                         & 224  & 40.9M  & 8.5    & \textbf{81.4}  & \textbf{95.6}  \\ \Xhline{2\arrayrulewidth}
\end{tabular}
\vspace{-0.26in}
\label{table:r1}
\end{table}

\begin{table}[!tb]\small
  \centering
  \caption{Performance comparisons with the state-of-the-art vision backbones for image recognition on ImageNet (advanced training setup). Models with similar top-1/top-5 accuracy are grouped for efficiency comparison.}
  \setlength{\tabcolsep}{0.5pt}
\begin{tabular}{c|c|cc|cc}
\Xhline{2\arrayrulewidth}
Backbone                                        & Res. & Params & GFLOPs & Top-1 Acc. & Top-5 Acc. \\ \hline
ResNet-50 \cite{he2016deep}                     & 224  & 25.5M  & 4.1    & 78.3       & 94.3       \\
CoaT-Lite Mini \cite{xu2021co}                  & 224  & 11M    & 2.0    & 78.9       & -          \\
EfficientNet-B1 \cite{tan2019efficientnet}      & 240  & 7.8M   & 0.7    & 79.1       & 94.4       \\
SE-ResNet-50 \cite{hu2018squeeze}               & 224  & 28.1M  & 4.1    & 79.4       & 94.6       \\
XCiT-T24 \cite{el2021xcit}                      & 224  & 12.1M  & 2.3    & 79.4       & -          \\
EfficientNet-B2 \cite{tan2019efficientnet}      & 260  & 9.2M   & 1.0    & 80.1       & 94.9       \\
BoTNet-S1-50 \cite{srinivas2021bottleneck}      & 224  & 20.8M  & 4.3    & 80.4       & 95.0       \\
ResNeSt-50-fast \cite{zhang2020resnest}         & 224  & 27.5M  & 4.3    & 80.6       & -          \\
ResNeSt-50      \cite{zhang2020resnest}         & 224  & 27.5M  & 5.4    & 81.1       & -          \\
Twins-PCPVT-S \cite{chu2021twins}               & 224  & 24.1M  & 3.7    & 81.2       & -          \\
Swin-T    \cite{liu2021swin}                    & 224  & 28.3M  & 4.5    & 81.3       & -          \\    \hline
\textbf{CoTNet-50}                              & 224  & 22.2M  & 3.3    & \textbf{81.3}  & \textbf{95.6}       \\
\textbf{CoTNeXt-50}                             & 224  & 30.1M  & 4.3    & \textbf{82.1}  & \textbf{95.9}       \\
\textbf{SE-CoTNetD-50}                          & 224  & 23.1M  & 4.1    & \textbf{81.6}  & \textbf{95.8}  \\ \hline\hline
ResNet-101 \cite{he2016deep}                    & 224  & 44.6M  & 7.9    & 80.0       & 95.0       \\
ResNet-152 \cite{he2016deep}                    & 224  & 60.2M  & 11.6   & 81.3       & 95.5       \\
SE-ResNet-101 \cite{hu2018squeeze}              & 224  & 49.3M  & 7.9    & 81.4       & 95.7       \\
TNT-S  \cite{han2021transformer}                & 224  & 23.8M  & 5.2    & 81.5       & 95.7       \\
EfficientNet-B3 \cite{tan2019efficientnet}      & 300  & 12.0M  & 1.8    & 81.6       & 95.7       \\
BoTNet-S1-59 \cite{srinivas2021bottleneck}      & 224  & 33.5M  & 7.3    & 81.7       & 95.8       \\
CoaT-Lite Small \cite{xu2021co}                 & 224  & 19.8M  & 4.0    & 81.9       & -          \\
ResNeSt-101-fast \cite{zhang2020resnest}        & 224  & 48.2M  & 8.1    & 82.0       & -          \\
ResNeSt-101 \cite{zhang2020resnest}             & 224  & 48.3M  & 10.2   & 82.3       & -          \\
LambdaResNet-101\cite{bello2021lambdanetworks}  & 224  & 36.9M  & -      & 82.3       & -          \\
XCiT-S24 \cite{el2021xcit}                      & 224  & 47.6M  & 9.1    & 82.6       & -          \\
CaiT-S-24 \cite{touvron2021going}               & 224  & 46.9M  & 9.4    & 82.7       & -           \\
Twins-PCPVT-B  \cite{chu2021twins}              & 224  & 56.0M  & 8.3    & 82.7       & -         \\   \hline
\textbf{CoTNet-101}                             & 224  & 38.3M  & 6.1    & \textbf{82.8}       & \textbf{96.2}       \\
\textbf{CoTNeXt-101}                            & 224  & 53.4M  & 8.2    & \textbf{83.2}       & \textbf{96.4}       \\
\textbf{SE-CoTNetD-101}                         & 224  & 40.9M  & 8.5    & \textbf{83.2}       & \textbf{96.5}  \\ \hline\hline
SE-ResNet-152 \cite{hu2018squeeze}              & 224  & 66.8M  & 11.6   & 82.2       & 95.9       \\
ConViT-B  \cite{d2021convit}                    & 224  & 86.5M  & 16.8   & 82.4       & 95.9       \\
BoTNet-S1-110 \cite{srinivas2021bottleneck}     & 224  & 54.7M  & 10.9   & 82.8       & 96.3       \\
TNT-B \cite{han2021transformer}                 & 224  & 65.6M  & 14.1   & 82.9       & 96.3       \\
XCiT-L24 \cite{el2021xcit}                      & 224  & 189.1M & 36.1   & 82.9       & -          \\
EfficientNet-B4 \cite{tan2019efficientnet}      & 380  & 19.0M  & 4.2    & 82.9       & 96.4       \\
CaiT-S-36 \cite{touvron2021going}               & 224  & 68.2M  & 13.9   & 83.3       & -           \\
Twins-PCPVT-L \cite{chu2021twins}               & 224  & 99.2M  & 14.8   & 83.3       & -          \\
Swin-B \cite{liu2021swin}                       & 224  & 87.7M  & 15.4   & 83.3       & -          \\
BoTNet-S1-128  \cite{srinivas2021bottleneck}    & 256  & 75.1M  & 19.3   & 83.5       & 96.5       \\
EfficientNet-B5 \cite{tan2019efficientnet}      & 456  & 30.0M  & 9.9    & 83.6       & 96.7       \\ \hline
\textbf{SE-CoTNetD-152}                         & 224  & 55.8M  & 17.0   & \textbf{84.0}       & \textbf{97.0}       \\ \hline\hline
SENet-350     \cite{hu2018squeeze}              & 384  & 115.2M & 52.9   & 83.8       & 96.6       \\
EfficientNet-B6 \cite{tan2019efficientnet}      & 528  & 43.0M  & 19.0   & 84.0       & 96.8       \\
BoTNet-S1-128 \cite{srinivas2021bottleneck}     & 320  & 75.1M  & 30.9   & 84.2       & 96.9       \\
Swin-B   \cite{liu2021swin}                     & 384  & 87.7M  & 47.0   & 84.2       & -          \\
EfficientNet-B7 \cite{tan2019efficientnet}      & 600  & 66.0M  & 37.0   & 84.3       & 97.0       \\ \hline
\textbf{SE-CoTNetD-152}                         & 320  & 55.8M  & 26.5   & \textbf{84.6}       & \textbf{97.1}       \\ \Xhline{2\arrayrulewidth}
\end{tabular}
\vspace{-0.26in}
\label{table:r2}
\end{table}

As shown in Table \ref{table:r1}, under the same depth (50-layer or 101-layer), the results across both top-1 and top-5 accuracy consistently indicate that our CoTNet-50/101 and CoTNeXt-50/101 obtain better performances against existing vision backbones with favorable parameter budget, including both ConvNets (e.g., ResNet-50/101 and ResNeXt-50/101) and attention-based models (e.g., Stand-Alone and AA-ResNet-50/101). The results generally highlight the key advantage of exploiting contextual information among keys in self-attention learning for visual recognition task. Specifically, under the same 50-layer backbones, by exploiting local self-attention in the deep architecture, LR-Net-50 and Stand-Alone exhibit better performance than ResNet-50, which ignores long-range feature interactions. Next, AA-ResNet-50 and LambdaResNet-50 enable the exploration of global self-attention over the whole feature map, and thereby boost up the performances. However, the performances of AA-ResNet-50 and LambdaResNet-50 are still lower than the stronger ConvNet (SE-ResNeXt-50) that strengthens the capacity of visual representation with channel-wise feature re-calibration. Furthermore, by fully replacing 3$\times$3 convolutions with CoT blocks across the entirety of deep architecture in ResNet-50/ResNeXt-50, CoTNet-50 and CoTNeXt-50 outperform SE-ResNeXt-50. This confirms that unifying both context mining among keys and self-attention learning into a single architecture is an effective way to enhance representation learning and thus boost visual recognition.
When additionally using exponential moving average as in LambdaResNet, the top-1 accuracy of CoTNeXt-50/101 will be further improved to 80.2\% and 81.3\% respectively, which is to-date the best published performance on ImageNet in default training setup.

Similar observations are also attained in advanced training setup, as summarized in Table \ref{table:r2}. Note that here we group all the baselines with similar top-1/top-5 accuracy or network depth. In general, our CoTNet-50 \& CoTNeXt-50 or CoTNet-101 \& CoTNeXt-101 perform consistently better than other vision backbones across both metrics for each group. In particular, the top-1 accuracy of our CoTNeXt-50 and CoTNeXt-101 can achieve 82.1\% and 83.2\%, making the absolute improvement over the best competitor ResNeSt-50 or ResNeSt-101/LambdaResNet-10 by 1.0\% and 0.9\%, respectively. More specifically, the attention-based backbones (BoTNet-S1-50 and BoTNet-S1-59) exhibit better performances than ResNet-50 and ResNet-101, by replacing the final three 3$\times$3 convolutions in ResNet with global self-attention layers. LambdaResNet-101 further boosts up the performances by leveraging the computationally efficient global self-attention layers (i.e., Lambda layer) to replace the convolutional layers. Nevertheless, LambdaResNet-101 is inferior to CoTNeXt-101 which capitalizes on the contextual information among input keys to guide self-attention learning. Even under the heavy setting with deeper networks, our SE-CoTNetD-152 (320) still manages to outperform the superior backbones of BoTNet-S1-128 (320) and EfficientNet-B7 (600), sharing the similar (even smaller) FLOPs with BoTNet-S1-128 (320).

\begin{figure}[!tb]
\vspace{-0.10in}
\centering {\includegraphics[width=0.4\textwidth]{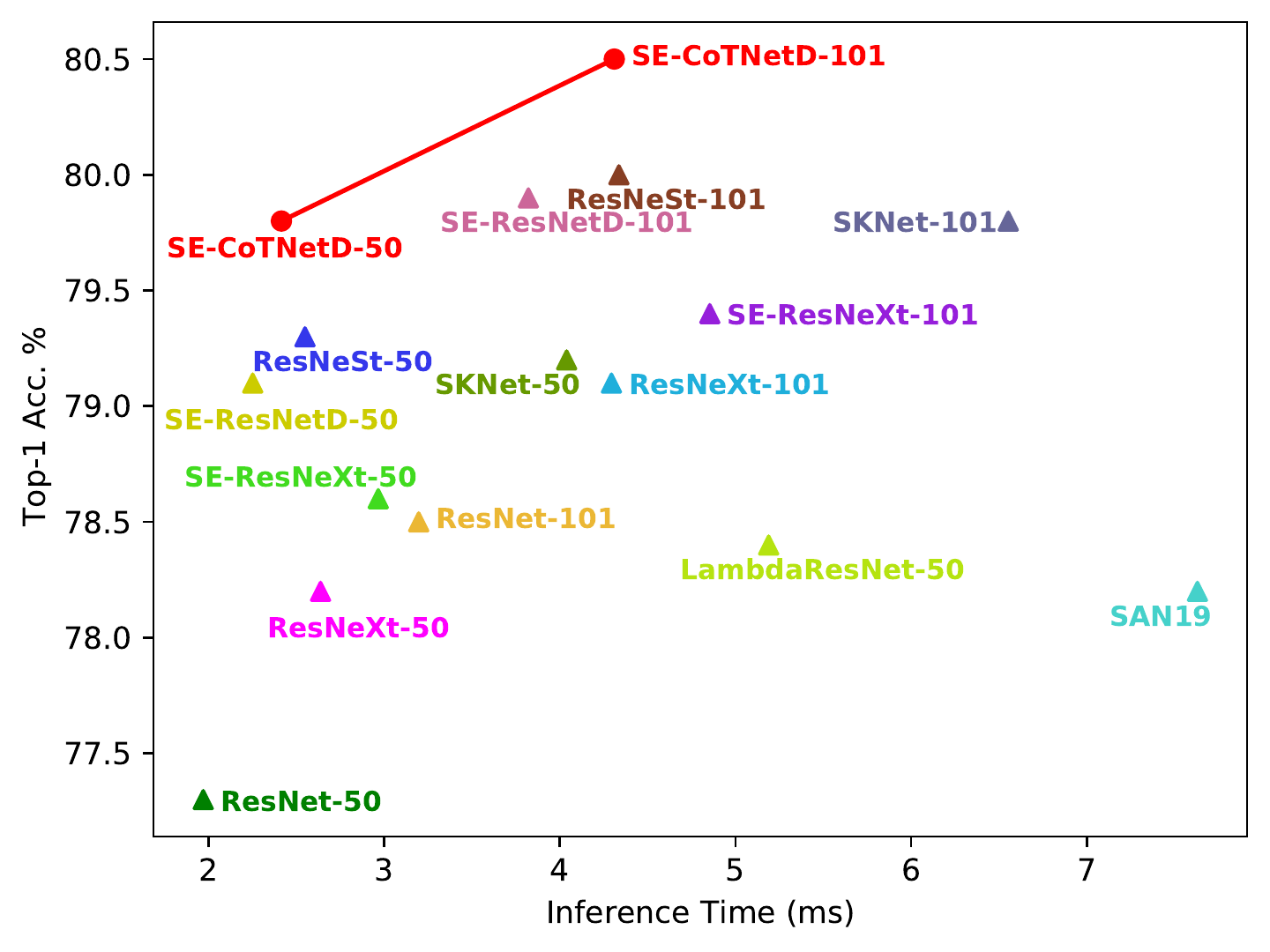}}
\vspace{-0.10in}
\caption{Inference Time vs. Accuracy Curve on ImageNet (default training setup).}
\vspace{-0.10in}
\label{fig:inference_time}
\end{figure}

\begin{figure}[!tb]
\vspace{-0.10in}
\centering {\includegraphics[width=0.4\textwidth]{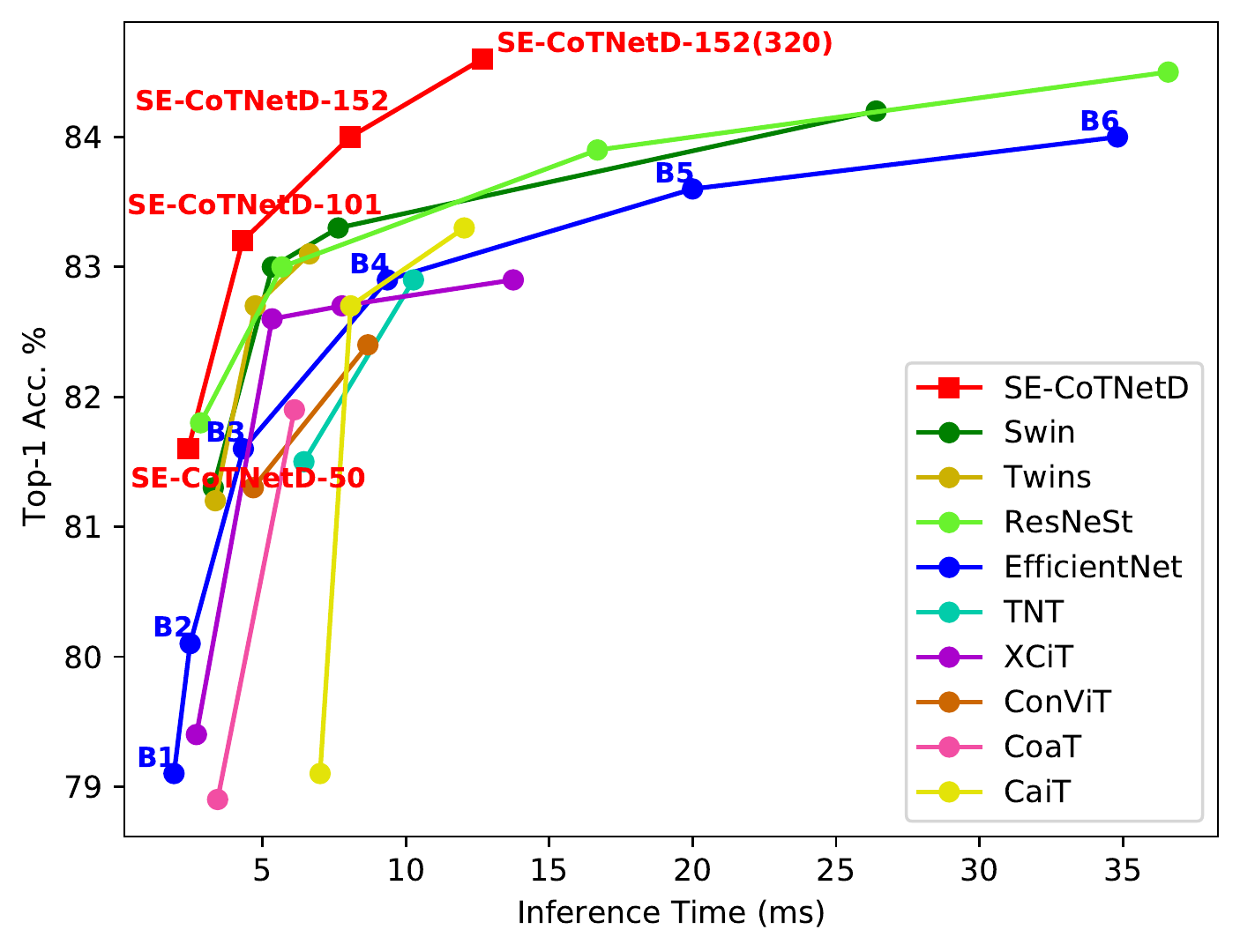}}
\vspace{-0.10in}
\caption{Inference Time vs. Accuracy Curve on ImageNet (advanced training setup).}
\vspace{-0.10in}
\label{fig:inference_time_advance}
\end{figure}

\begin{table}[!tb]\small
  \centering
  \caption{Performance comparisons across different ways on the exploration of contextual information, i.e., using only static context (\textbf{Static Context}), using only dynamic context (\textbf{Dynamic Context}), linearly fusing static and dynamic contexts (\textbf{Linear Fusion}), and the full version of CoT block. The backbone is CoTNet-50 and we adopt the default setup for training on ImageNet.}
  \setlength{\tabcolsep}{3.5pt}
\begin{tabular}{c|c|c|c|c}
\Xhline{2\arrayrulewidth}
                & Params & GFLOPs & Top-1 Acc. & Top-5 Acc. \\ \hline
Static Context     & 17.1M  & 2.7    & 77.1       & 93.5          \\
Dynamic Context     & 20.3M  & 3.3    & 78.5       & 94.1       \\
Linear Fusion   & 20.3M  & 3.3    & 78.7       & 94.2       \\ \hline
CoT             & 22.2M  & 3.3    & 79.2       & 94.5       \\ \Xhline{2\arrayrulewidth}
\end{tabular}
\vspace{-0.22in}
\label{table:as}
\end{table}

\textbf{Inference Time vs. Accuracy.}
Here we evaluate our CoTNet models with regard to both inference time and top-1 accuracy for image recognition task. Figure \ref{fig:inference_time} and Figure \ref{fig:inference_time_advance} show the inference time-accuracy curve under both default and advanced training setups for our CoTNet and the state-of-the-art vision backbones. As shown in the two figures, we can see that our CoTNet models consistently obtain better top-1 accuracy with less inference time than other vision backbones across both training setups. In a word, our CoTNet models seek better inference time-accuracy trade-offs than existing vision backbones. More remarkably, compared to the high-quality backbone of EfficientNet-B6, our SE-CoTNetD-152 (320) achieves 0.6\% higher top-1 accuracy, while runs 2.75$\times$ faster at inference.

\begin{table*}[!tb]\small
  \centering
  \caption{Effect of utilizing different replacement settings on the four stages (\textbf{res2}$\rightarrow$\textbf{res3}$\rightarrow$\textbf{res4}$\rightarrow$\textbf{res5}) in the basic backbone of ResNet-50 and two widely adopted architecture changes, ResNet-D \cite{he2019bag} and Squeeze-and-Excitation \cite{hu2018squeeze} (\textbf{D-SE}). $\checkmark$ denotes the stage is replaced with our CoT blocks. $\star$ denotes the use of architecture changes (D-SE). We adopt the default setup for training on ImageNet.}
  \setlength{\tabcolsep}{3.8pt}
\begin{tabular}{c|ccccc|c|c|c|c|c}
\Xhline{2\arrayrulewidth}
               & res2         & res3         & res4          & res5          & D-SE          & Params & GFLOPs & Infer  & Top-1 Acc. & Top-5 Acc.   \\ \hline
  ResNet-50    &              &              &               &               &               & 25.5M  & 4.1    & 508 ex/s  & 77.3       & 93.6     \\ \hline
  \multirow{4}{*}{CoTNet-50}
               &              &              &               & $\checkmark$  &               & 23.5M  & 4.0    & 491 ex/s  & 78.5       & 94.1         \\
               &              &              & $\checkmark$  & $\checkmark$  &               & 22.4M  & 3.7    & 443 ex/s  & 79.0       & 94.3         \\
               &              & $\checkmark$ & $\checkmark$  & $\checkmark$  &               & 22.3M  & 3.4    & 390 ex/s  & 79.0       & 94.4         \\
               & $\checkmark$ & $\checkmark$ & $\checkmark$  & $\checkmark$  &               & 22.2M  & 3.3    & 331 ex/s  & 79.2       & 94.5 \\\hline
  SE-ResNetD-50 &             &              &               &               & $\star$       & 35.7M  & 4.4    & 444 ex/s  & 79.1       & 94.5          \\
  SE-CoTNetD-50 &             &              & $\checkmark$  & $\checkmark$  & $\star$       & 23.1M  & 4.1    & 414 ex/s  & 79.8       & 94.7          \\
\Xhline{2\arrayrulewidth}
\end{tabular}
\vspace{-0.22in}
\label{table:rs}
\end{table*}

\textbf{Ablation Study.}
In this section, we investigate how each design in our CoT block influences the overall performance of CoTNet-50. In CoT block, we first mine the static context among keys via a 3$\times$3 convolution. Conditioned on the concatenation of query and contextualized key, we can also obtain the dynamic context via self-attention. CoT block dynamically fuses the static and dynamic contexts as the final outputs. Here we include one variant of CoT block by directly summating the two kinds of contexts, named as Linear Fusion.

Table \ref{table:as} details the performances across different ways on the exploration of contextual information in CoTNet-50 backbone. Solely using static context (Static Context) for image recognition achieves 77.1\% top-1 accuracy, which can be interpreted as one kind of ConvNet without self-attention. Next, by directly exploiting the dynamic context via self-attention, Dynamic Context exhibits better performance. The linear fusion of static and dynamic contexts leads to a boost of 78.7\%, which basically validates the complementarity of the two contexts. CoT block is further benefited from the dynamic fusion via attention, and the top-1 accuracy of CoT finally reaches 79.2\%.

\textbf{Effect of Replacement Settings.}
In order to show the relationship between performance and the number of stages replaced with our CoT blocks, we progressively replace the stages with our CoT blocks in ResNet-50 backbone (res2$\rightarrow$res3$\rightarrow$res4$\rightarrow$res5), and compare the performances. The results shown in Table \ref{table:rs} indicate that increasing the number of stages replaced with CoT blocks can generally lead to performance improvement, and meanwhile the parameter number \& FLOPs are slightly decreased. When taking a close look on the throughputs and accuracies of different replacement settings, the replacement of CoT blocks in the last two stages (res4 and res5) contributes to the most performance boost. The additional replacement of CoT blocks in the fist stages (res1 and res2) can only lead to a marginal performance improvement (0.2\% top-1 accuracy in total), while requiring 1.34$\times$ inference time. Therefore, in order to seek a better speed-accuracy trade-off, we follow \cite{bello2021lambdanetworks} and construct an upgraded version of our CoTNet, named SE-CoTNetD-50, where only the 3$\times$3 convolutions in the res4 and res5 stages are replaced with CoT blocks under SE-ResNetD-50 backbone. Note that the SE-ResNetD-50 backbone is a variant of ResNet-50 with two widely adopted architecture changes (ResNet-D \cite{he2019bag} and Squeeze-and-Excitation in all bottleneck blocks \cite{hu2018squeeze}). As shown in Table \ref{table:rs}, compared to the SE-ResNetD-50 counterpart, our SE-CoTNetD-50 achieves better performances at a virtually negligible decrease in throughput.

\subsection{Object Detection}

\textbf{Setup.} We next evaluate the pre-trained CoTNet for the downstream task of object detection on COCO dataset. For this task, we adopt Faster-RCNN \cite{ren2016faster,ren2015faster} and Cascade-RCNN \cite{cai2018cascade} as the base object detectors, and directly replace the vanilla ResNet backbone with our CoTNet. Following the standard setting in \cite{xie2017aggregated}, we train all models on COCO-2017 training set ($\sim$118K images) and evaluate them on COCO-2017 validation set (5K images). The standard AP metric of single scale is adopted for evaluation. During training, for each input image, the size of the shorter side is sampled from the range of [640, 800]. All models are trained with FPN \cite{lin2017feature} and synchronized batch normalization \cite{zhang2018context}. We utilize the 1x learning rate schedule for training. For fair comparison with other vision backbones in this task, we set all the hyperparameters and detection heads as in \cite{zhang2020resnest}.

\textbf{Performance Comparison.} Table \ref{table:od} summarizes the performance comparisons on COCO dataset for object detection with Faster-RCNN and Cascade-RCNN in different pre-trained backbones. We group the vision backbones with same network depth (50-layer/101-layer). From observation, our pre-trained CoTNet models (CoTNet-50/101 and CoTNeXt-50/101) exhibit a clear performance boost against the ConvNets backbones (ResNet-50/101 and ResNeSt-50/101) for each network depth across all IoU thresholds and object sizes. The results basically demonstrate the advantage of integrating self-attention learning with contextual information mining in CoTNet, even when transferred to the downstream task of object detection.

\begin{table}[!tb]\scriptsize
  \centering
  \caption{Performance comparisons with the state-of-the-art vision backbones on the downstream task of object detection (Base detectors: Faster-RCNN and Cascade-RCNN). Average Precision (AP) is reported at different IoU thresholds and for three different object sizes: small, medium, large (s/m/l).}
  \setlength{\tabcolsep}{2.5pt}
\begin{tabular}{c|c|ccc|ccc}
\Xhline{2\arrayrulewidth}
  & Backbone                      & $AP$    & $AP_{50}$ & $AP_{75}$ & ${AP_s}$ & $AP_m$ & $AP_{l}$    \\ \hline
\multirow{10}{*}{\rotatebox[origin=c]{90}{Faster-RCNN}}
        & ResNet-50 \cite{he2016deep}         & 39.34   & 59.47     & 42.76     & 23.57    & 42.42  & 51.30       \\
        & ResNeXt-50 \cite{xie2017aggregated} & 41.31   & 62.23     & 44.91     & 25.33    & 44.52  & 53.20      \\
        & ResNeSt-50 \cite{zhang2020resnest}  & 42.39  & 63.73     & 46.02     & 26.25    & 45.88  & 54.24       \\
        & CoTNet-50   & \textbf{43.50}  & \textbf{64.84}     & \textbf{47.53}     & \textbf{26.36}   & \textbf{47.54}  & \textbf{56.49}   \\
        & CoTNeXt-50  & \textbf{44.06}  & \textbf{65.76}     & \textbf{47.65}     & \textbf{27.08}   & \textbf{47.70}  & \textbf{57.21}  \\ \cline{2-8}
        & ResNet-101 \cite{he2016deep}        & 41.46  & 61.99     & 45.38     & 25.31    & 44.75  & 54.62       \\
        & ResNeXt-101 \cite{xie2017aggregated} & 42.91 & 63.77     & 46.89     & 25.96    & 46.42  & 55.47       \\
        & ResNeSt-101 \cite{zhang2020resnest} & 44.13  & 61.91     & 47.67     & 26.02    & 47.69  & 57.48       \\
        & CoTNet-101  & \textbf{45.35}  & \textbf{66.80}     & \textbf{49.18}  & \textbf{28.65}    & \textbf{49.47}  & \textbf{58.82}       \\
        & CoTNeXt-101 & \textbf{46.10}  & \textbf{67.50}     & \textbf{50.22}  & \textbf{29.44}    & \textbf{49.84}  & \textbf{59.26} \\ \hline\hline
\multirow{10}{*}{\rotatebox[origin=c]{90}{Cascade-RCNN}}
        & ResNet-50 \cite{he2016deep}         & 42.45  & 59.76     & 46.09     & 24.90    & 45.64  & 55.86       \\
        & ResNeXt-50 \cite{xie2017aggregated} & 44.53  & 62.45     & 48.38     & 27.29    & 48.01  & 57.87       \\
        & ResNeSt-50 \cite{zhang2020resnest}  & 45.41  & 63.92     & 48.70     & 28.77    & 48.69  & 58.43      \\
        & CoTNet-50   & \textbf{46.11}  & \textbf{64.68}     & \textbf{49.75}  & 28.71    & \textbf{49.76}  & \textbf{60.28}       \\
        & CoTNeXt-50  & \textbf{46.79}  & \textbf{65.54}     & \textbf{50.53}  & \textbf{29.74}    & \textbf{50.49}  & \textbf{61.04}       \\ \cline{2-8}
        & ResNet-101 \cite{he2016deep}        & 44.13  & 61.91     & 47.67     & 26.02    & 47.69  & 57.48       \\
        & ResNeXt-101 \cite{xie2017aggregated} & 45.83 & 63.61     & 49.89     & 27.75    & 49.53  & 59.14       \\
        & ResNeSt-101 \cite{zhang2020resnest} & 47.51  & 66.06     & 51.35     & 30.25    & 50.96  & 61.23       \\
        & CoTNet-101  & \textbf{48.19}  & \textbf{67.00}     & \textbf{52.17}  & 30.00    & \textbf{52.32}  & \textbf{62.87}       \\
        & CoTNeXt-101 & \textbf{49.02}  & \textbf{67.67}     & \textbf{53.03}  & \textbf{31.44}    & \textbf{52.95}  & \textbf{63.17} \\
\Xhline{2\arrayrulewidth}
\end{tabular}
\vspace{-0.22in}
\label{table:od}
\end{table}

\subsection{Instance Segmentation}

\textbf{Setup.} Here we evaluate the pre-trained CoTNet in another downstream task of instance segmentation on COCO dataset. This task goes beyond the box-level understanding in object detection by additionally predicting the object mask for each detected object, pursuing the pixel-level understanding of visual content. Specifically, Mask-RCNN \cite{he2017mask,he2018mask} and Cascade-Mask-RCNN \cite{cai2018cascade} are utilized as the base models for instance segmentation. In the experiments, we replace the vanilla ResNet backbone in Mask-RCNN with our CoTNet. Similarly, all models are trained with FPN and synchronized batch normalization. We adopt the 1x learning rate schedule during training, and all the other hyperparameters are set as in \cite{zhang2020resnest}. For evaluation, we report the standard COCO metrics including both bounding box and mask AP (${AP}^{bb}$ and ${AP}^{mk}$).

\begin{table}[!tb]\scriptsize
  \centering
  \caption{Performance comparisons with the state-of-the-art vision backbones on the downstream task of instance segmentation (Base models: Mask-RCNN and Cascade-Mask-RCNN). The bounding box and mask Average Precision (${AP}^{bb}$, ${AP}^{mk}$) are reported at different IoU thresholds. Note that BoTNet-50/101 is fine-tuned with larger input size 1024$\times$1024 and longer epochs (36).}
  \setlength{\tabcolsep}{1.8pt}
\begin{tabular}{c|c|ccc|ccc}
\Xhline{2\arrayrulewidth}
 & Backbone                                 & $AP^{bb}$ & $AP^{bb}_{50}$ & $AP^{bb}_{75}$ & $AP^{mk}$ & $AP^{mk}_{50}$ & $AP^{mk}_{75}$ \\ \hline
 \multirow{12}{*}{\rotatebox[origin=c]{90}{Mask-RCNN}}
    & ResNet-50 \cite{he2016deep}              & 39.97     & 60.19          & 43.73          & 36.05     & 57.02          & 38.54          \\
    & ResNeXt-50 \cite{xie2017aggregated}      & 41.74     & 62.32          & 45.60          & 37.41     & 59.24          & 39.98          \\
    & ResNeSt-50 \cite{zhang2020resnest}       & 42.81     & 63.93          & 46.85          & 38.14     & 60.54          & 40.69          \\
    & BoTNet-50 \cite{srinivas2021bottleneck}  & 43.6      & 65.3           & 47.6           & 38.9      & 62.5           & 41.3           \\
    & CoTNet-50   & \textbf{44.06}             & 64.99   & \textbf{48.29}    & \textbf{39.28}     & 62.12          & \textbf{42.17}          \\
    & CoTNeXt-50  & \textbf{44.47}    & \textbf{65.74} & \textbf{48.71}    & \textbf{39.62}     & \textbf{62.70}      & \textbf{42.35} \\ \cline{2-8}
    & ResNet-101 \cite{he2016deep}             & 41.78     & 61.90          & 45.80          & 37.50     & 58.78          & 40.21          \\
    & ResNeXt-101 \cite{xie2017aggregated}     & 43.25     & 63.61          & 47.23          & 38.60     & 60.74          & 41.37          \\
    & ResNeSt-101 \cite{zhang2020resnest}      & 45.75     & 66.88          & 49.75          & 40.65     & 63.76          & 43.68          \\
    & BoTNet-101 \cite{srinivas2021bottleneck} & 45.5      & -              & -              & 40.4      &  -            &  -            \\
    & CoTNet-101          & \textbf{46.17}  & \textbf{67.17}  & \textbf{50.63}  & \textbf{40.86}  & \textbf{64.18}   & 43.64        \\
    & CoTNeXt-101         & \textbf{46.66}  & \textbf{67.70}  & \textbf{50.90}  & \textbf{41.21}  & \textbf{64.45}  & \textbf{44.27} \\ \hline\hline
\multirow{10}{*}{\rotatebox[origin=c]{90}{Cascade-Mask-RCNN}}
        & ResNet-50 \cite{he2016deep}         & 43.06  & 60.29     & 46.55     & 37.19    & 57.61  & 40.01       \\
        & ResNeXt-50 \cite{xie2017aggregated} & 44.91  & 62.66     & 48.80     & 38.57    & 59.83  & 41.59       \\
        & ResNeSt-50 \cite{zhang2020resnest}  & 46.23  & 64.62     & 50.15     & 39.64    & 61.86  & 42.88       \\
        & CoTNet-50   & \textbf{46.94}  & \textbf{65.36}     & \textbf{50.69}  & \textbf{40.25}    & \textbf{62.37}  & \textbf{43.38}       \\
        & CoTNeXt-50  & \textbf{47.63}  & \textbf{65.93}     & \textbf{51.64}  & \textbf{40.76}    & \textbf{63.32}  & \textbf{44.01}       \\ \cline{2-8}
        & ResNet-101 \cite{he2016deep}         & 44.79  & 62.31     & 48.46     & 38.51    & 59.33  & 41.53       \\
        & ResNeXt-101 \cite{xie2017aggregated} & 46.24  & 64.01     & 49.92     & 39.77    & 61.19  & 43.06       \\
        & ResNeSt-101 \cite{zhang2020resnest}  & 48.44  & 66.80     & 52.60     & 41.52    & 64.03  & 45.02       \\
        & CoTNet-101  & \textbf{48.97}  & \textbf{67.42}     & \textbf{53.10}  & \textbf{41.98}    & \textbf{64.81}  & \textbf{45.39}       \\
        & CoTNeXt-101 & \textbf{49.35}  & \textbf{67.88}     & \textbf{53.53}  & \textbf{42.20}    & \textbf{65.00}  & \textbf{45.69} \\
     \Xhline{2\arrayrulewidth}
\end{tabular}
\vspace{-0.22in}
\label{table:is}
\end{table}

\textbf{Performance Comparison.}
Table \ref{table:is} details the performances of Mask-RCNN with different pre-trained vision backbones for the downstream task of instance segmentation on COCO dataset. Similar to the observations for object detection downstream task, our pre-trained CoTNet models yields consistent gains against both ConvNets backbones (ResNet-50/101 and ResNeSt-50/101) and attention-based model (BoTNet-50/101) over the most IoU thresholds. This generally highlights the generalizability of our CoTNet in the challenging instance segmentation task. In particular, BoTNet-50 achieves better performances than the best ConvNets (ResNeSt-50). This might attribute to the additional modeling of global self-attention in BoTNet plus the more advanced fine-tuning setup with larger input size (1024$\times$1024) and longer training epochs (36). However, by uniquely exploiting the contextual information among neighbor keys for self-attention learning, our CoTNet-50 manages to lead the performance boosts over the most metrics, even when fine-tuned with smaller input size and less epoches (12). The results again confirm the merit of simultaneously performing context mining and self-attention learning in our CoTNet for visual representation learning.

\section{Conclusions}

In this work, we propose a new Transformer-style architecture, termed Contextual Transformer (CoT) block, which exploits the contextual information among input keys to guide self-attention learning. CoT block first captures the static context among neighbor keys, which is further leveraged to trigger self-attention that mines the dynamic context. Such way elegantly unifies context mining and self-attention learning into a single architecture, thereby strengthening the capacity of visual representation. Our CoT block can readily replace standard convolutions in existing ResNet architectures, meanwhile retaining the favorable parameter budget. To verify our claim, we construct Contextual Transformer Networks (CoTNet) by replacing the 3$\times$3 convolutions in ResNet architectures (e.g., ResNet or ResNeXt). The CoTNet architectures learnt on ImageNet validate our proposal and analysis. Experiments conducted on COCO in the context of object detection and instance segmentation also demonstrate the generalization of the visual representation pre-trained by our CoTNet.

{\small
\bibliographystyle{ieee_fullname}
\bibliography{egbib}
}

\end{document}